\documentclass{article}
\usepackage{spconf,amsmath,graphicx,amsfonts,amssymb,amsopn}
\usepackage{subfig} 
\usepackage{float}
\graphicspath{{Figures/}}

\def\qunit{\bv{\hat{q}}}
\newcommand{\figref}[1]{Fig.\,\ref{#1}}
\newcommand{\bv}[1]{\boldsymbol{#1}}
\title{An Optimal Dimensionality Multi-shell Sampling Scheme with Accurate and Efficient Transforms for Diffusion MRI}
 \name{Alice P. Bates$^{\star}$ \qquad Zubair Khalid$^{\star}$ \qquad Jason D. McEwen$^{\dagger}$ \qquad Rodney A. Kennedy$^{\star}$}

 \address{$^{\star}$ Research School of Engineering, The Australian National University, Canberra, Australia \\
     $^{\dagger}$ Mullard Space Science Laboratory, University College London, Surrey RH5 6NT, UK}

\begin{document}
%
\maketitle
\begin{abstract}
This paper proposes a multi-shell sampling scheme and corresponding transforms for the accurate reconstruction of the diffusion signal in diffusion MRI by expansion in the spherical polar Fourier (SPF) basis. The sampling scheme uses an optimal number of samples, equal to the degrees of freedom of the band-limited diffusion signal in the SPF domain, and allows for computationally efficient reconstruction. We use synthetic data sets to demonstrate that the proposed scheme allows for greater reconstruction accuracy of the diffusion signal than the multi-shell sampling scheme obtained using the generalised electrostatic energy minimisation (gEEM) method used in the Human Connectome Project. We also demonstrate that the proposed sampling scheme allows for increased angular discrimination and improved rotational invariance of reconstruction accuracy than the gEEM scheme.
\end{abstract}

\section{Introduction} \label{sec:intro}
In diffusion MRI, the intra-voxel diffusion of water molecules is used to determine the structure and connectivity of white matter in the brain. The diffusion signal can be reconstructed from a finite number of measurements in $q$-space, where $\bv{q}$ is the diffusion wave vector. Diffusion signal measurements are normally collected on a single sphere, if only
angular information is required, or multiple concentric spheres in $q$-space, known as $q$-shells, if both angular and radial information is required~\cite{assemlal:2009b}. The diffusion signal can then be reconstructed by expansion in an orthonormal basis, such as the spherical polar Fourier (SPF) basis~\cite{assemlal:2009b}. Diffusion characteristics, such as the ensemble average propagator (EAP) and its features,
such as the orientation distribution function (ODF), can be obtained analytically using the SPF coefficients~\cite{assemlal:2009b,Cheng2010}.

The number of diffusion signal measurements that can be acquired must be small due to the need for scan times to be practical in a clinical setting. In addition, for accurate and fast reconstruction of the diffusion signal and subsequent computation of the diffusion characteristics, the SPF coefficients must be computed accurately and efficiently. The minimum number
of samples required by any sampling scheme to allow accurate reconstruction of the diffusion signal by expansion in the SPF basis is equal to the degrees of freedom of the signal in the
SPF basis, hence we referred to this as the optimal number of samples~\cite{Bates:2015c,Bates:2015b}. Furthermore,
as within each voxel white-matter fibre populations may assume any orientation, the diffusion signal reconstruction accuracy should not change significantly if the diffusion signal, or equivalently the sampling scheme, is rotated~\cite{caruyer:2013,cheng:2014}.

Several multi-shell sampling schemes are proposed in the literature, the majority of these focus on uniform sampling of the sphere to achieve rotationally invariant reconstruction accuracy
and aim to maximise the angular distribution of samples within and between shells to achieve increased angular discrimination~[5-7]. Most of these schemes distributed the shell radii uniformly~[5-7]. Existing multi-shell sampling schemes use least-squares to calculate the diffusion signal coefficients, which is computationally intensive~[5-8]. 

 Electrostatic energy minimisation and spherical code are two methods for obtaining a uniform arrangement of samples, these methods have been generalised to multi-shell sampling schemes in~\cite{caruyer:2013} and \cite{cheng:2014} respectively. \cite{caruyer:2013,cheng:2014} use a cost function that is a trade-off of sample location uniformity within a shell compared with the uniformity of all samples projected onto one sphere to obtain uniformity within and between shells. The electrostatic energy minimisation scheme \cite{caruyer:2013} can be used to design incremental acquisition schemes.

The multi-shell sampling scheme proposed in~\cite{ye:2012} uses the concept of uniform and dual polyhedra with alternating shells having complementary sets of directions. However, this scheme only provides two separate sets of directions, which only allows different directions on each shell for sampling schemes with two shells. Also, due to the limited number of uniform polyhedra, this scheme cannot be used to construct sampling grids of arbitrary size. In \cite{alipoor:2015} icosahedral sampling schemes with arbitrary
number of samples are developed; this work has not been generalised to multi-shell sampling.

A sampling scheme that increases reconstruction accuracy by minimising the condition number of the least-squares matrix was developed in~\cite{caruyer:2012b}. This scheme does not have exact quadrature
as the number of samples per shell is proportional to the Gauss-Laguerre weights which are not integers. The scheme also has more than the optimal number of samples.

We here develop a sampling scheme and corresponding transform for reconstructing the diffusion signal in the SPF basis. The proposed multi-shell framework attains an optimal number of samples, equal to the degrees of freedom of the diffusion signal band-limited in the SPF basis, and has an efficient transform for calculating the coefficients in the SPF basis. We demonstrate that the
proposed sampling scheme allows for accurate and effectively rotationally invariant reconstruction of the
diffusion signal, as well as allowing for greater angular discrimination compared with the generalised electrostatic energy minimisation sampling scheme~\cite{caruyer:2013}.

\section{Materials and Methods} \label{sec:mat_and_methods}

\subsection{SPF Basis}
The diffusion signal, specifically the normalised MR signal attenuation, $E(\bv{q})$ can be expanded in the SPF basis \cite{assemlal:2009b} as
\begin{equation}
\label{Eq:E_expansion}
    E(\bv{q})=\sum_{n=0}^{N-1}\sum_{\ell=0}^{L-1}\sum_{m=-{\ell}}^{\ell} (E)_{n\ell m}  R_n(q)Y_{\ell}^m(\qunit),
\end{equation}
where $\qunit = \frac{\bv{q}}{|\bv{q}|}$, $q = |\bv{q}|$,  $Y_{\ell}^m(\qunit)$ is the spherical harmonic coefficient of degree $\ell$ and order $m$, and the expansion coefficients are given by
\begin{equation}\label{Eq:Ecoeff}
(E)_{n\ell m} = \langle E(\bv{q}),  R_n(q)Y_{\ell}^m(\qunit)\rangle.
\end{equation}
The radial functions $R_n$ are Gaussian-Laguerre polynomials.
\begin{equation}
\label{Eq:radial_func}
  R_n(q)=\bigg[\frac{2}{\zeta^{1.5}}\frac{n!}{\Gamma(n+1.5)}\bigg]^{0.5}\exp\bigg(\frac{-q^2}{2\zeta}\bigg)L_n^{1/2}\bigg(\frac{q^2}{\zeta}\bigg),
\end{equation}
%
where $\zeta$ denotes the scale factor and $L_n^{1/2}$ are the $n$-th generalised Laguerre polynomials of order half. The expansion in (\ref{Eq:E_expansion}) assumes that $E(\bv{q})$ is band-limited at radial order $N$ and angular order $L$.

\subsection{Proposed Sampling Scheme and Transform}
Due to the separability of the SPF basis, the 3D transform for calculating the diffusion signal coefficients \eqref{Eq:Ecoeff} can be separated into transforms in the radial and angular directions.

For the radial transformation, Gauss-Laguerre quadrature can be used, where $N$ sampling nodes is sufficient for exact quadrature. The $N$ shells of the proposed multi-shell sampling scheme are placed at $q_i = \sqrt{\zeta x_i}$ where $x_i$ are the roots of the $N$-th generalised Laguerre polynomial of order a half. We determine the corresponding weights to be
\begin{equation}
\label{Eq:weight}
w_i =\frac{0.5\zeta^{1.5}\Gamma(N+1.5)x_ie^{x_i}}{N!(N+1)^2[L_{N+1}^{0.5}(x_i)]^2}.
\end{equation}

It was found in \cite{assemlal:2009b} that $N=4$ shells are sufficient for convergence to the ground truth when the signal was Gaussian or bi-Gaussian.  We set the scaling factor $\zeta$ so that shells are located at $b$-values within an interval of interest. In this work, we use a maximum $b$-value of 8000 $\rm{s/mm}^2$, as in \cite{daducci:2011}, resulting in shells at $b = 411.3, 1694.4, 4036.3$ and $8000$ $\rm{s/mm}^2$.

For sampling within each shell, we use the recently proposed single-shell sampling scheme~\cite{Bates:2015c,Bates:2015b} which allows accurate reconstruction, with the reconstruction error on the order of machine precision for signals band-limited in the spherical harmonic basis, has an efficient forward and inverse spherical harmonic transforms (SHT), and uses an optimal number of samples for the band-limited diffusion signal on the sphere, equal to $L(L+1)/2$.

The spherical harmonic band-limit, and therefore the number of samples in each shell, is determined using \cite{daducci:2011}, where the authors determined the spherical harmonic band-limit $L$ required to accurately reconstruct the Gaussian diffusion signal at different $b$-values, the shells have $L =$ 3, 5, 9 and 11 for the inner most (smaller $b$-value) to outer most shell respectively. The proposed sampling scheme therefore has a total of 132 samples.

\begin{figure}[t]
  \centering
  \subfloat[]{
  \includegraphics[width=0.22\textwidth]{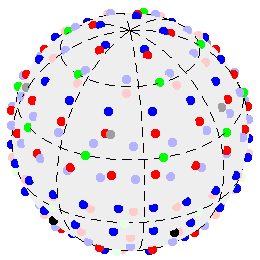}}
   \subfloat[]{
  \includegraphics[width=0.22\textwidth]{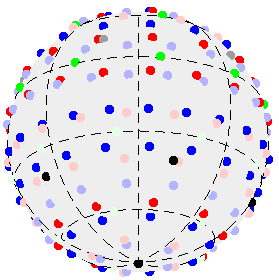}}
  \caption{Proposed sampling scheme (a) North pole view and (b) South pole view.}
  \label{fig:sampling_grid}
\end{figure}

\figref{fig:sampling_grid} shows the proposed sampling scheme projected onto a single sphere, samples on the inner most to outer most shell are shown in black, green, red and blue for each shell respectively. Locations where antipodal symmetry is used to infer the value of the signal are lighter in color.

\vspace{-2mm}
\subsubsection{Computational Complexity}
The proposed transform for computing the SPF coefficients is composed of $N$ SHTs, where $N$ is the number of shells. Using the SHT proposed in \cite{Bates:2015b}, for the case of all
shells having the same spherical harmonic band-limit $L$, the proposed sampling scheme has computational complexity
$O(NL^4)$ which is significantly faster than least-squares method of computation which has computational complexity  $O(NL^6)$. As the inner shells have smaller band-limits than the outer shells,
the computational complexity is actually smaller than $O(NL^4)$.

\section{Results} \label{sec:pagestyle}
In order to evaluate the multi-shell sampling grid and corresponding transform presented in Section 2, we compare with the generalised electrostatic energy minimisation (gEEM) method~\footnote{Available at https://github.com/ecaruyer/qspace}~\cite{caruyer:2013} used in the Human Connectome Project which has shells evenly distributed radially, we set $q_{max}$ and $q_{min}$ and the number of samples per shell to be the same as the proposed scheme. The non-incremental version of the gEEM scheme is used for fair comparison. When the gEEM sampling scheme is used with the optimal number of samples, regularisation of the least-squares matrix is required as the matrix is ill-conditioned; we use $\lambda_l = 10^{-7}$ and $\lambda_n = 5 \times 10^{-8}$ for the angular and radial regularisation respectively as in \cite{Cheng2010}.

\subsection{Reconstruction Accuracy}
We evaluate the reconstructed accuracy of the proposed scheme using three different Gaussian models of the diffusion signal: one fibre (Gaussian diffusion), and two fibres (mixture of Gaussians) with a $90^\circ$ and $45^\circ$ crossing angle. The Gaussian models have diffusivities of $\lambda_1=1.7$ mm$^2/$s and $\lambda_2=\lambda_3=0.2$ mm$^2/$s. For each model of the diffusion signal, synthetic data sets are generated using the proposed sampling scheme presented in Section 2. The coefficients of the diffusion signal in the SPF basis $(E)_{n\ell m}$ are then calculated using the proposed transform and finally the diffusion signal is reconstructed using \eqref{Eq:E_expansion}.

\begin{figure}[!t]
    \centering
    \includegraphics[clip = true, trim = 0 0mm 0 7mm, width=0.48\textwidth]{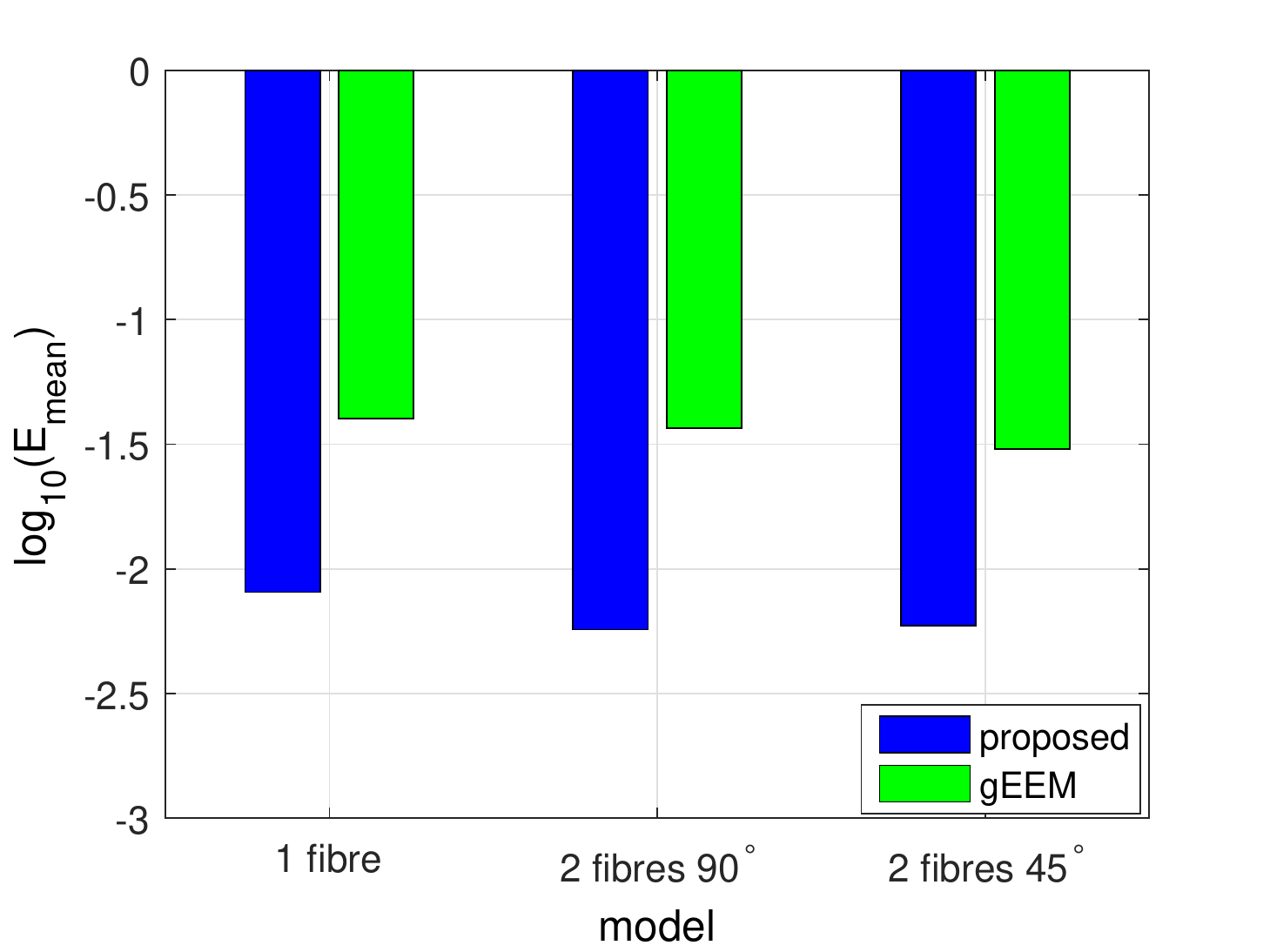}
    \caption{Log of the mean reconstruction error $\rm{log}_{10}(E_{\rm{mean}})$ for the gEEM and the proposed sample scheme for three different Gaussian models of the diffusion signal.}
    \label{fig:mean_error}
\end{figure}

The reconstruction error between the reconstructed signal $E_R(\bv{q})$ and the ground truth obtained from each model $E_M(\bv{q})$ is evaluated on a Cartesian sampling grid (samples
uniformly distributed in $q$-space) with 10000 samples and the mean error, $E_{\rm{mean}} \triangleq \sum_{i=1}^{10000}(|E_M(\bv{q_i})- E_R(\bv{q_i})  |)/10000$ is calculated. \figref{fig:mean_error} shows log$_{10}(E_{\rm{mean}})$ for all three diffusion signal models. It is evident that the proposed scheme has a smaller reconstruction error for all three models than the gEEM sampling scheme. It was found that proposed sampling scheme had a smaller reconstruction error throughout $q$-space, particularly at large $q$-space radii. Due to space constraints, these results are not shown here.

\subsection{Rotational Invariance of Reconstruction Accuracy}
\begin{figure}[!t]
    \centering
    \includegraphics[clip = true, trim = 0 0mm 0 7mm, width=0.48\textwidth]{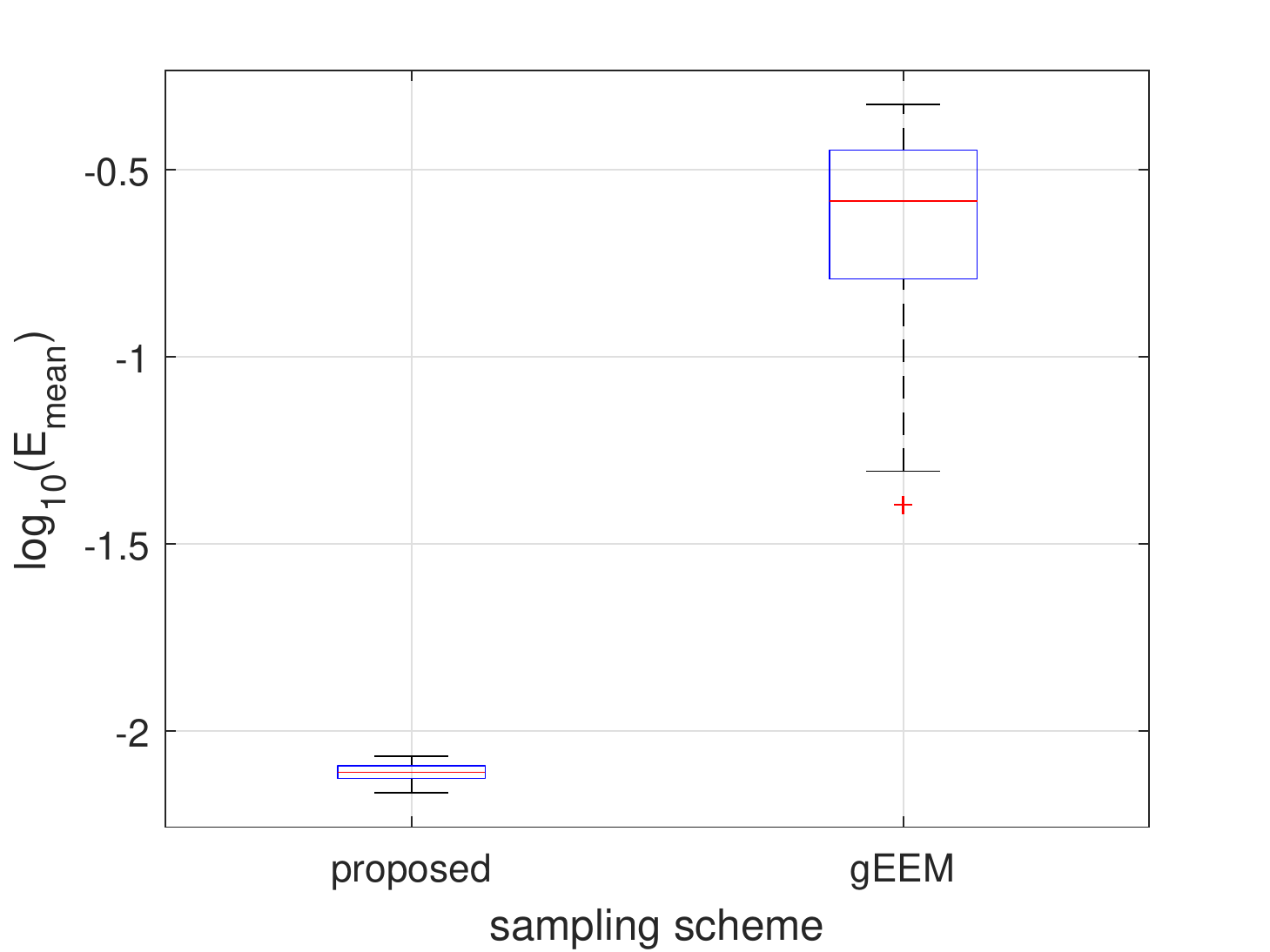}
    \caption{Log of the mean reconstruction error $\rm{log}_{10}(E_{\rm{mean}})$ for different rotations of the single fibre (Gaussian) model of the diffusion signal for the gEEM scheme and the proposed sample scheme.}
    \label{fig:rot}
\end{figure}

We analyse how the reconstruction accuracy depends on fibre orientation by calculating $E_{\rm{mean}}$ for 30 randomly generated rotations of the diffusion signal model with one fibre (Gaussian diffusion). The log of the mean reconstruction error $\rm{log}_{10}(E_{\rm{mean}})$ is shown in \figref{fig:rot}. The proposed
sampling scheme shows smaller variation in reconstruction error compared with the gEEM scheme.

\subsection{Angular Discrimination}
\begin{figure}[!t]
    \centering
    \includegraphics[clip = true, trim = 0 0mm 0 7mm, width=0.48\textwidth]{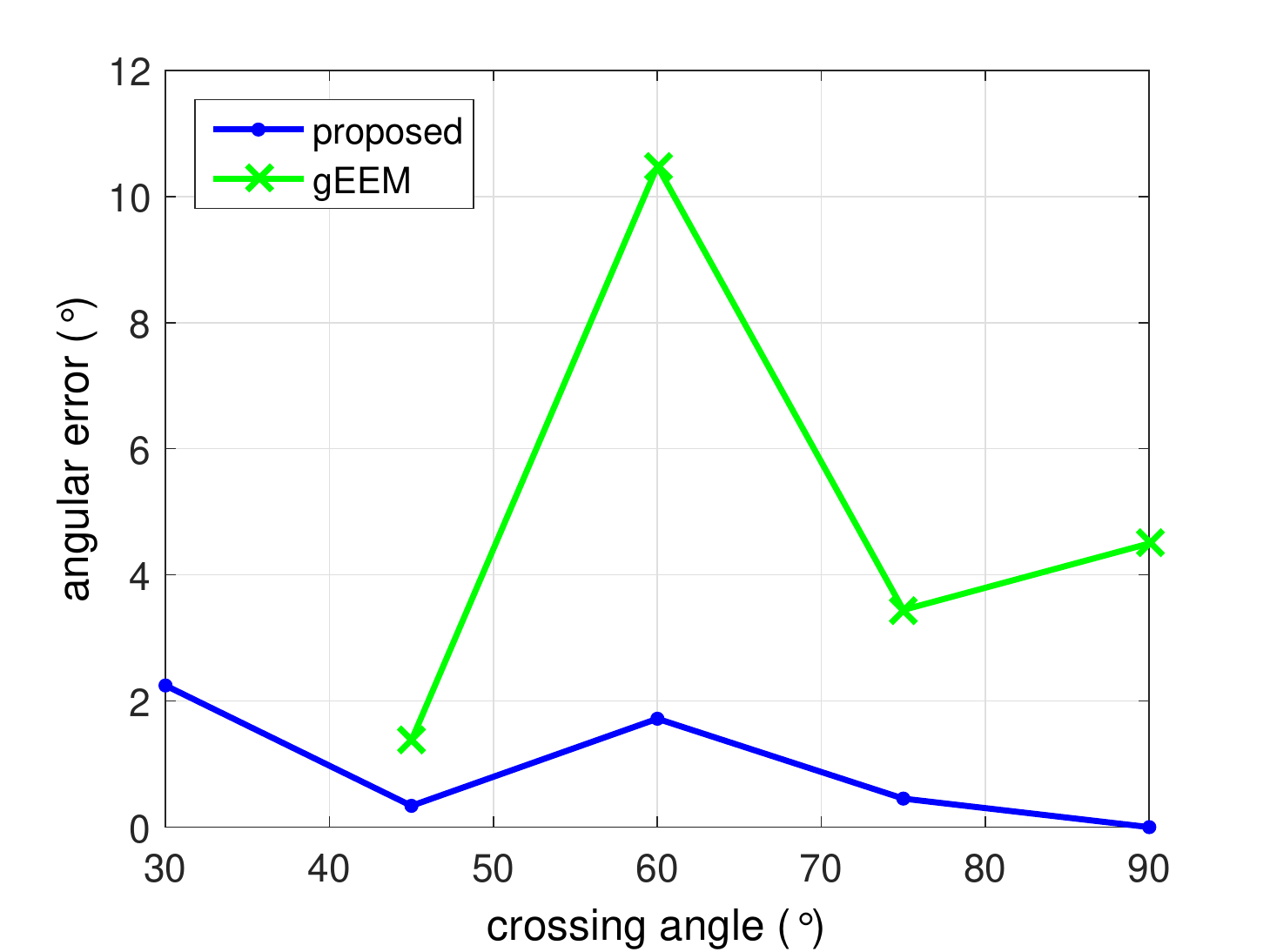}
    \caption{Angular error for fibre crossing angles between $30^\circ$ and $90^\circ$ for the gEEM and the proposed sample scheme.}
    \label{fig:ang_error}
\end{figure}

In order to evaluate the proposed sampling scheme in terms of angular discrimination, we use the analytical expression in~\cite{Cheng2010} to obtaining the ODF from the SPF coefficients.
We find fibre directions by a discrete search for the maxima of the ODF over 2562 vertices on the sphere. \figref{fig:ang_error} shows the mean angular error for the proposed and gEEM
sampling scheme for two fibres (mixture of Gaussians) at crossing angles between $30^\circ$ to $90^\circ$. For all crossing angles the proposed scheme has a smaller angular error.
The gEEM sampling scheme is unable to resolve two fibres for the $30^\circ$ crossing angle, also for some crossing angles more than two fibre populations were incorrectly detected.

\section{Discussion and Conclusions}
We have proposed a novel multi-shell sampling scheme for the measurement and reconstruction of the diffusion signal. The scheme has an optimal number of samples equal to the degrees of freedom of the band-limited diffusion signal in the SPF basis and is computationally efficient. We evaluated the proposed scheme using synthetic data sets by comparing its performance with that of the gEEM scheme used in the Human Connectome Project. The least-squares method of reconstruction used by the gEEM method is ill-conditioned with an optimal number of samples. While regularisation improves this, the proposed transform still results in greater reconstruction accuracy, leading to more rotationally invariant reconstruction accuracy and improved angular discrimination.

As future work, we intend to apply our sampling scheme to more complicated models of diffusion and consider the effect of noise on reconstruction accuracy, before using our scheme for real image acquisitions.

\section{Acknowledgments} This work is supported by the Australian Research Council's Discovery Projects funding scheme (Project no. DP170101897). The authors would like to thank Dr. Emmanuel Caruyer for making his code publicly available and providing assistance for generating the gEEM sampling scheme used as a benchmark in this paper.

\bibliographystyle{IEEEbib}
\bibliography{multi-shell_database}

\end{document}